\documentclass[runningheads]{llncs}

\usepackage{eccv}

\usepackage{eccvabbrv}

\usepackage{graphicx}
\usepackage{booktabs}
\usepackage{multirow}        
\usepackage{setspace}
\usepackage{float}
\usepackage{mathtools}
\usepackage{algpseudocode}
\usepackage{amsmath}
\RequirePackage{times}    % Integrate Times for here
\RequirePackage{cite}     % Automatically ordered citations
\RequirePackage{xspace}
\RequirePackage{amsmath}
\RequirePackage{amssymb}
\RequirePackage[ruled,linesnumbered]{algorithm2e}

\usepackage[accsupp]{axessibility}  % Improves PDF readability for those with disabilities.

\usepackage{hyperref}

\usepackage{orcidlink}

\begin{document}

\title{Temporal Residual Guided Diffusion Framework for Event-Driven Video Reconstruction} 

%Event-driven Video Reconstruction: A Temporal Residual Guided Diffusion Framework
%A Temporal Residual Guided Diffusion Framework for Event-driven Video Reconstruction
\titlerunning{Temporal Residual Guided Diffusion Framework}

\author{Lin Zhu\inst{1}\orcidlink{0000-0001-6487-0441} \and
Yunlong Zheng\inst{1}\orcidlink{0009-0008-1882-3124} \and
Yijun Zhang\inst{2}\orcidlink{0000-0003-2289-2372} \and
Xiao Wang\inst{3}\orcidlink{0000-0001-6117-6745} \and
Lizhi Wang\inst{1}\orcidlink{0000-0002-1953-3339} \and
Hua Huang\inst{4,}\thanks{Corresponding author.}\orcidlink{0000-0003-2587-1702}}

\authorrunning{Lin Zhu et al.}

\institute{Beijing Institute of Technology, Beijing, China \and
China Mobile (Suzhou) Software Technology Co., Ltd., Jiangsu, China \and
Anhui University, Anhui, China \and
Beijing Normal University, Beijing, China}

\maketitle

\begin{abstract}
Event-based video reconstruction has garnered increasing attention due to its advantages, such as high dynamic range and rapid motion capture capabilities. However, current methods often prioritize the extraction of temporal information from continuous event flow, leading to an overemphasis on low-frequency texture features in the scene, resulting in over-smoothing and blurry artifacts. Addressing this challenge necessitates the integration of conditional information, encompassing temporal features, low-frequency texture, and high-frequency events, to guide the Denoising Diffusion Probabilistic Model (DDPM) in producing accurate and natural outputs.
To tackle this issue, we introduce a novel approach, the Temporal Residual Guided Diffusion Framework, which effectively leverages both temporal and frequency-based event priors. Our framework incorporates three key conditioning modules: a pre-trained low-frequency intensity estimation module, a temporal recurrent encoder module, and an attention-based high-frequency prior enhancement module. 
In order to capture temporal scene variations from the events at the current moment, we employ a temporal-domain residual image as the target for the diffusion model.
Through the combination of these three conditioning paths and the temporal residual framework, our framework excels in reconstructing high-quality videos from event flow, mitigating issues such as artifacts and over-smoothing commonly observed in previous approaches. Extensive experiments conducted on multiple benchmark datasets validate the superior performance of our framework compared to prior event-based reconstruction methods. %\textcolor{red}{Our code is available at \url{https://github.com/BIT-Vision/TRDF-E2VID}.}
\end{abstract}

\section{Introduction}
\label{sec:intro}

\begin{figure}
    \centering
    \includegraphics[width=0.7\linewidth, angle=0]{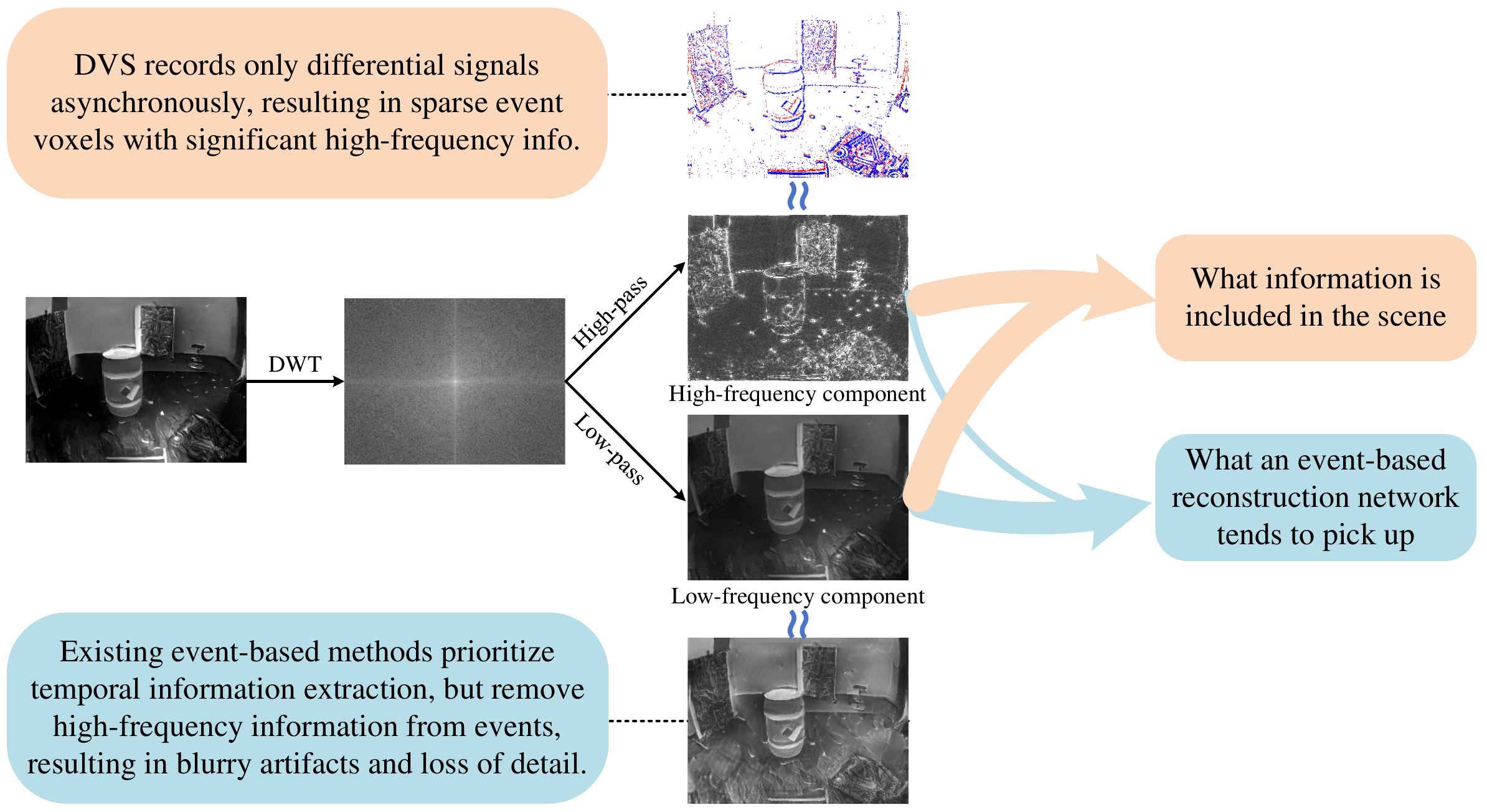}
    \caption{Existing methods often emphasize low-frequency texture, causing over-smoothing and loss of high-frequency details in image reconstruction. This motivated us to explore a framework that strategically incorporates both temporal and high-frequency event priors.}
    \label{fig:cmpfrequency}
\end{figure}

Event-based video reconstruction has emerged as a prominent area of research, fueled by the unique advantages offered by event cameras, such as high dynamic range and rapid motion capture capabilities~\cite{Sensor1, Sensor2}. However, existing methods~\cite{E2VID, E2VID-TPAMI, FireNet, SPADE, zhu2021neuspike, ETNet, zhu2022event} tend to prioritize temporal information extraction from continuous event flow, leading to an overemphasis on low-frequency texture features and resulting in issues like over-smoothing and artifacts in reconstructed scenes. It is crucial to recognize the unique nature of event cameras, which operate differently from conventional cameras by measuring high-frequency changes in intensity asynchronously at the time they occur.
However, despite their potential, the temporal nature of real event cameras complicates the reconstruction problem~\cite{gu2023reliable,xiang2022temporal}, necessitating innovative solutions.
Fig. \ref{fig:cmpfrequency} illustrates a comparison between the frequency domain components of the scene and the reconstruction results of event-based reconstruction methods.

Recently, Diffusion Probabilistic Models (DPMs)~\cite{DDPM, NCSN, DALLE, ILVR, Stable, DiffusionCLIP,zeng2024improving} have shown remarkable progress in image generation, opening avenues for event-based reconstruction tasks by using conditional DPMs to generate realistic images with fine details. The integration of an initial predictor and a diffusion model ensures precise and adaptive conditioning, resulting in perceptual quality improvements of restored images. However, challenges persist in effectively integrating degraded images and other conditional information into DPMs for improved generative capacity. Incorporating various streams of conditional information into clusters of DPMs is essential for facilitating spatiotemporal adaptive conditioning throughout the reconstruction process. To address this challenge, we propose to integrate conditional information, encompassing temporal, low-frequency texture, and high-frequency events, to guide the Denoising Diffusion Probabilistic Model (DDPM) in generating accurate and natural outputs.

In this paper, we introduce a novel approach named the temporal residual guided diffusion framework. This framework strategically leverages both temporal and frequency-based event priors and incorporates three key conditioning modules: a pre-trained low-frequency intensity estimation module, a temporal recurrent encoder module, and an attention-based high-frequency prior enhancement module. To capture temporal scene variations effectively, we introduce the use of a temporal-domain residual image as the target for the diffusion model. By combining these three conditioning paths with the temporal residual framework, our proposed framework excels in reconstructing high-quality videos from event flow, addressing issues such as artifacts and over-smoothing commonly observed in previous approaches. In summary, this paper introduces innovative frameworks in the realms of event-based video reconstruction and diffusion-based image restoration. Our proposed temporal residual guided diffusion video framework and unified conditional framework for image restoration significantly advance the state-of-the-art, addressing key challenges and showcasing superior performance through extensive experiments. The contributions of our paper can be summarized as follows:

1) We introduce a novel temporal residual guided diffusion framework. This framework combines temporal and frequency-based event priors to effectively capture temporal scene variations in video reconstruction.

2) The framework strategically incorporates three conditioning modules, namely a low-frequency intensity estimation module, a temporal recurrent encoder module, and an attention-based high-frequency prior enhancement module. Through the amalgamation of these conditioning paths with the temporal residual framework, our proposed approach excels in reconstructing high-quality videos from event flow.

3) Extensive experiments demonstrate that our framework is effective in overcoming issues like artifacts and over-smoothing, establishing itself as a noteworthy advancement in both event-based video reconstruction and diffusion-based image restoration.

\section{Related Works}

\noindent\textbf{Event-based Video Reconstruction.}
In virtue of its high dynamic range, high temporal resolution, and low power consumption, event sensors excel in many visual application scenarios.
Video reconstruction is a fundamental and popular topic in the event-based vision literature.
Early event-based video reconstruction approaches relied on the representational similarity between events and gradients~\cite{SMosaic, IMFVI, RIRMR} or optical flow~\cite{OFIE} to reconstruct scene structure directly. Nevertheless, these early methods fell short in achieving sufficiently realistic final reconstructed intensity images, primarily due to a lack of exploration into the prior information embedded in long-term data.

Since the widespread adoption of deep learning methods, many data-driven approaches have demonstrated significant potential.
E2VID~\cite{E2VID, E2VID-TPAMI} utilizes LSTM to accumulate temporal features of events and learns on large datasets with optical flow constraints, resulting in significant improvements in reconstruction quality.
FireNet~\cite{FireNet} employs a lighter network architecture to achieve faster reconstruction speeds.
Sparse-E2VID~\cite{sparse-e2vid} only emphasizes high-frequency by learning gradients with lightweight networks and then utilizes these gradients to generate images following traditional reconstruction approaches.
{Sparse-E2VID~\cite{sparse-e2vid} emphasizes high-frequency by learning gradients with lightweight networks and then utilizes these gradients to generate images following traditional reconstruction approaches.}
SPADE-E2VID~\cite{SPADE} uses previously reconstructed images to conditionally modulate the activations on a layer, significantly improving the effectiveness of sparse event reconstruction.
ETNet~\cite{ETNet} presents a hybrid CNN-Transformer~\cite{VIT} structure to reconstruct video, achieving the best reconstruction results to date.
However, these methods do not effectively balance the relationship between long-term and short-term event features, resulting in reconstructed images often containing motion artifacts and blur.

\noindent\textbf{Conditional Diffusion Model in Image/video Restoration.}
The process from intensity images to events can be modeled by a degradation model dominated by a differential operator.
Therefore, event-based video reconstruction task can be viewed as a class of video restoration tasks, and the prior relevant work has greatly benefited our research.
In the wake of DDPM~\cite{DDPM} demonstrating the impressive capability of diffusion models to generate images from random noise, applying diffusion model to image restoration tasks has become a popular focus in the literature.
SR3~\cite{SR3} and Platte~\cite{palette} use degraded images as conditions, directly concatenating them with noisy images as the overall input to the network, demonstrating the compatibility and efficiency of diffusion model for image restoration.
ShadowDiffusion~\cite{ShadowDiffusion}, IDM~\cite{IDM}, and DeS3~\cite{DeS3} utilize preprocessing features from existing models as conditions for the diffusion model.
Resdiff~\cite{Resdiff} and UCDIR~\cite{UCDIR} employ additional preprocessing models to roughly restore degraded images before utilizing conditional diffusion models to generate residuals, significantly reducing the processing complexity of the task.
IR-SDE~\cite{IR-SDE} and Refusion~\cite{Refusion} alter the diffusion process of DDPM itself, allowing sampling to commence from noisy degraded images rather than pure random noise, which significantly reduces both training and sampling costs.
Inspired by multimodal tasks, Refusion~\cite{Refusion} and DiffIR~\cite{DiffIR} encode images into a latent space to perform diffusion processes, accelerating the process of single-shot training and sampling.
Event-Diffusion~\cite{EventDiffusion} attempts to apply the diffusion model to an event-based image reconstruction task with the input of reconstructed images and original events.

The various methods mentioned above involve directly retraining the diffusion model to handle image restoration tasks, which can be computationally expensive. However, there is a class of methods that leverage a pre-trained diffusion model and can handle image restoration tasks at an extremely low cost.
SNIPS~\cite{SNIPS}, DDRM~\cite{DDRM}, and DDNM~\cite{DDNM} employ well-known degradation models to perform decomposition, making full use of the denoising prior of the diffusion model.
Inspired by the use of classifier gradients in DMBG~\cite{DMBG} for image synthesis, DPS~\cite{DPS} and GDP~\cite{GDP} calculate the posterior distribution based on Bayesian theory to guide the mean of each step in the sampling process.

In this paper, we introduce a novel temporal residual guided diffusion video framework, which integrates conditional information, encompassing temporal low-frequency texture and high-frequency events, to guide DDPM in event-based video reconstruction.

\section{Methodology}

\subsection{Problem Statement} 

Event-based reconstruction tasks aim to reconstruct images from a series of events $\mathcal{E}$:
\begin{equation}
    \mathcal{E} = \{e^i\}^{N-1}_{i=0} = \{(x^i, y^i, t^i, p^i)\}^{N-1}_{i=0},
\end{equation}
where $x^i$, $y^i$ represent the pixel positions of the event $e^i$, and $t^i$ represents the timestamp. Additionally, $p^i$ represents the polarity of $e^i$, indicating the relationship between the scene brightness and time variation, which can be calculated as follows:
\begin{equation}
	p^i = \begin{cases}
	+1, \ \ \  L(x,y,t^i) - L(x,y,t^i-\delta t) \geq \varphi_{+}\\
	-1, \ \ \  L(x,y,t^i) - L(x,y,t^i-\delta t) \leq \varphi_{-}
		   \end{cases},
\end{equation}
where $L$ represents the scene intensity in the logarithmic domain, $\delta t$ is the time interval from the previous event occurrence, and $\varphi_{+}, \varphi_{-}$ respectively denote the intensity thresholds for positive and negative polarities.

In a manner similar to the application of voxel-based organization in ~\cite{UEOF}, we also transform events into two-dimensional images along the temporal channel with $B$ bins:
\begin{equation}
    V^t(b) = \Sigma_{i=1}^{N} p_i \max(0, 1-|b-\frac{t^i-t^0}{t^{N-1}-t^0}(B-1)|).
\end{equation}

Considering sparse events contain more high-frequency information, we segment events between two frames based on event density, ultimately ensuring that the event density for each processing step does not exceed 0.25.
Then, we can further transform the problem we need to address: Given a set of voxel grids $\{V^t\}^{T-1}_{t=0}$, estimate the scene video corresponding to each moment $\{\tilde{I}^t\}^{T-1}_{t=0}$.
\begin{figure*}[t]
    \centering
    \includegraphics[width=0.9\linewidth, angle=0]{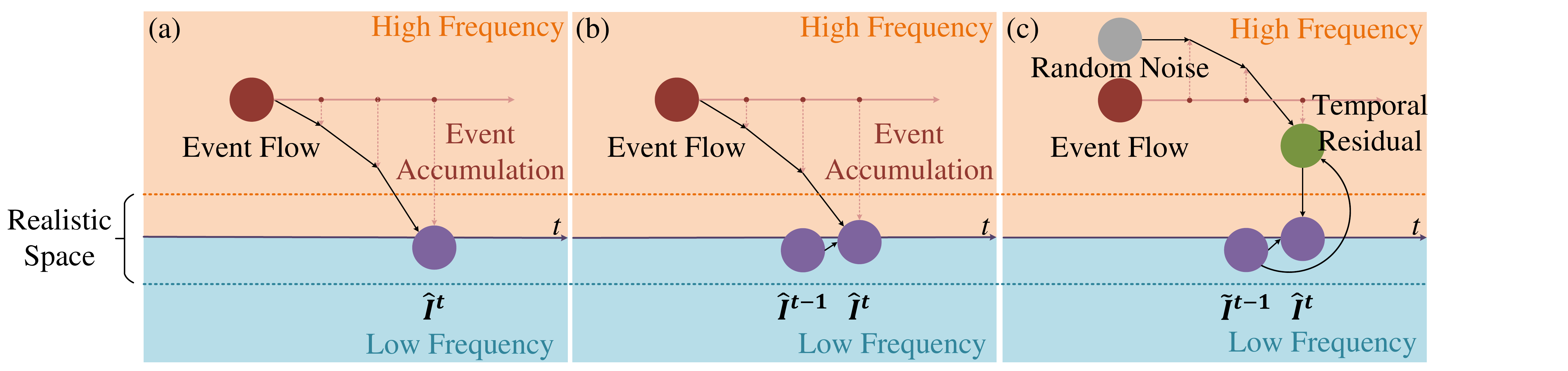}
    %\vspace{-6mm}
    \caption{Comparison of different strategies. (a) Directly predict intensity images from the accumulation features of past events, such as E2VID~\cite{E2VID, E2VID-TPAMI}, ETNet~\cite{ETNet}. (b) Jointly reconstruction from event feature accumulations and prediction from the previous frame, e.g., SPADE-E2VID~\cite{SPADE}. (c) Our temporal residual guided diffusion framework. While most methods adopt the initial two strategies, the inherent temporal feature extracting in these approaches results in the forfeiture of high-frequency information from the events. Our approach effectively tackles this contradiction by generating high-frequency temporal residuals through a conditional diffusion model.}%  Please see Sec.~\ref{sec:32} for a detailed analysis.}
    %\vspace{-3mm}
    \label{fig:frequency}
\end{figure*}

\subsection{Frequency-based Event Priors Analysis}\label{sec:32}

Fig.~\ref{fig:frequency} illustrates the current mainstream strategies for event-based reconstruction.
The reconstructed image becomes more realistic with a more balanced representation of low-frequency and high-frequency information.
Most methods~\cite{E2VID, FireNet, ETNet} adopt the approach corresponding to (a) in Fig.~\ref{fig:frequency}, directly accumulating event information in the time domain to estimate scene intensity.
This is extremely challenging, and they can achieve different approximations of realistic quality based on different architectures and algorithm complexities.
SPADE-E2VID~\cite{SPADE} adopts the approach corresponding to (b) in Fig.~\ref{fig:frequency}, building upon the predictions from the previous frame to simplify the task and achieve more realistic images with less computational cost.
Our method, as illustrated by (c) in Fig.~\ref{fig:frequency}, is based on the initial intensity estimation from the previous frame.
It incorporates certain low-frequency details while generating high-frequency temporal residual images through high-frequency random noise and events at the current time.
Therefore, our algorithm is capable of achieving more realistic reconstruction results.

\begin{figure*}[t]
    \centering
    \includegraphics[width=0.8\linewidth, angle=0]{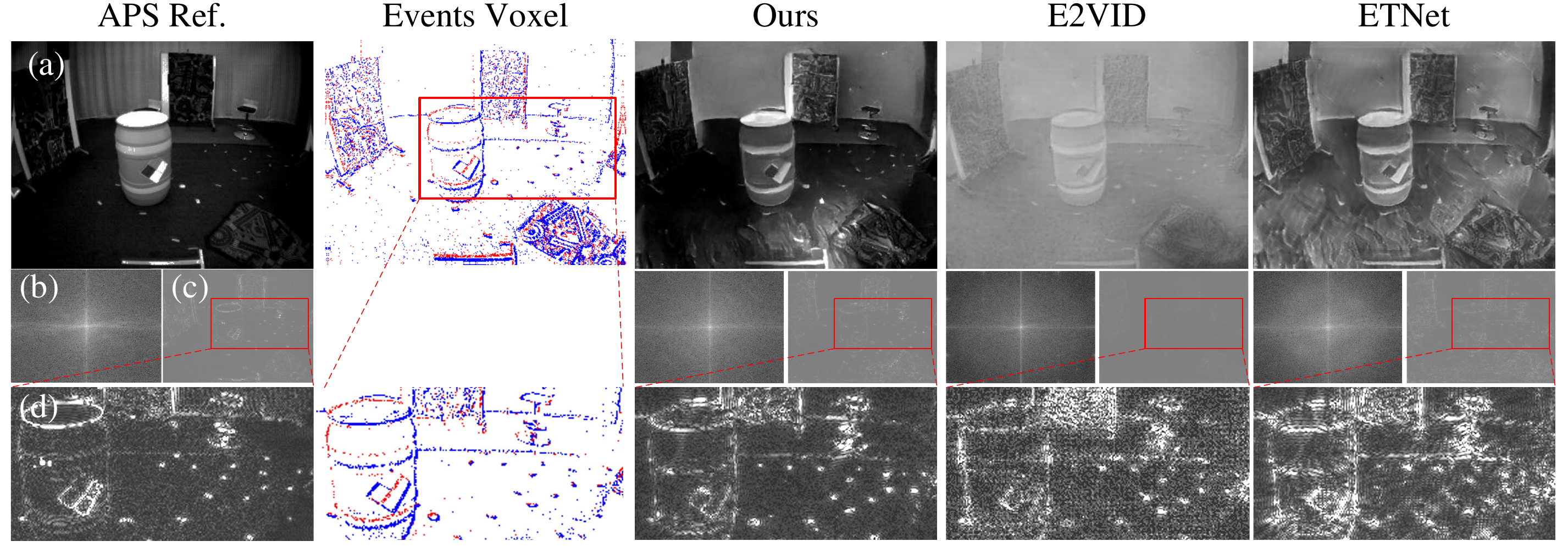}
    %\vspace{-3mm}
    \caption{Frequency domain analysis of reconstructed results. (a) Original image; b) Fourier spectrum chart of intensity image; (c) High-frequency components of intensity image; (d) Local magnification diagram (scaled for representation). Despite the events being very similar to the high-frequency map of the scene, E2VID and ETNet cannot reconstruct precise high-frequency details.}
    \label{fig:cmp_frequency}
\end{figure*}

We perform frequency domain analysis on the reconstructed results of different methods, as shown in Fig.~\ref{fig:cmp_frequency}.
Due to the limitations of the approach, both ETNet~\cite{ETNet} and E2VID~\cite{E2VID} produce results with a noticeable blur at the edges, along with a reduction in high-frequency components and increased susceptibility to noise interference.
Instead, the high-frequency components of our results closely resemble the events voxel, demonstrating that our method effectively utilizes the high-frequency features of events at the current time.
% \subsection{Overview of Conditional Denoising Framework}\label{sec:CDF}

\subsection{Temporal Residual Diffusion Framework}\label{sec:CDF}

\begin{figure*}[t]
    \centering
    \includegraphics[width=1\linewidth, angle=0]{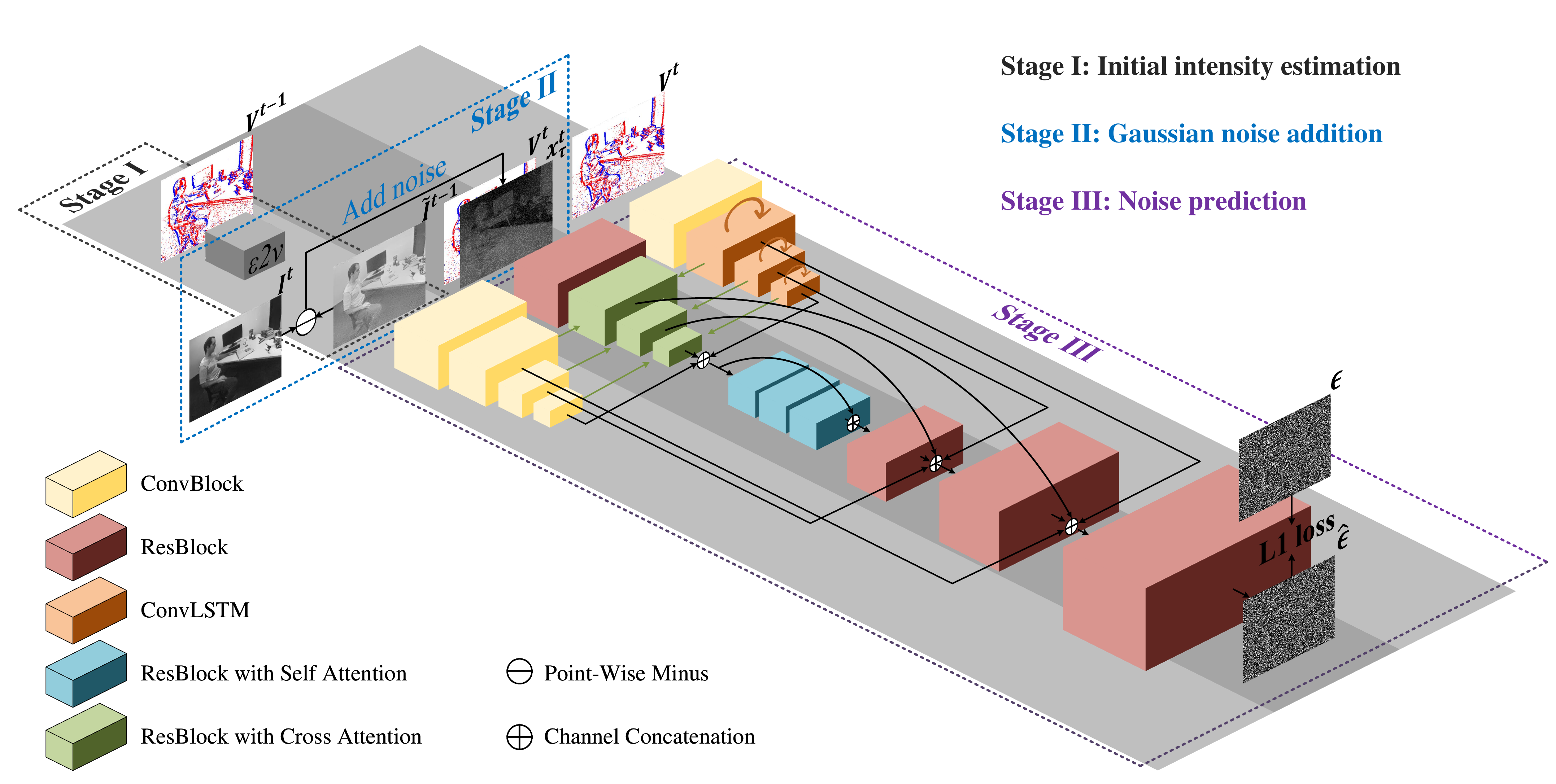}
    %\vspace{-3mm}
    \caption{Overview of temporal residual diffusion framework. At Stage I, a pre-trained intensity predictor generates initial low-frequency estimation; At Stage II, the residual image is computed in the time domain and noise is added; At Stage III, a triple-path conditional model is used to predict noise. Please refer to Fig.~\ref{fig:samplingAca} for specific details on the ResBlock with Cross Attention.}
    %\vspace{-3mm}
    \label{fig:training}
\end{figure*}

Fig. \ref{fig:training} shows the training phase of the proposed temporal residual diffusion framework.
Event data represents the change in intensity over time and constitutes a differential signal. In contrast, intensity images depict the normalized brightness of the scene and constitute an integral signal.
Due to their different modalities, there is a gap in data distribution, so using event data as a condition for directly generating intensity images is theoretically suboptimal, which has been confirmed in our experiments as well.
To simplify the generation task and harness the temporal variations in the scene represented by the event voxel grid at the current moment, we use temporal-domain residual image $x^t_0$ as the target images for the diffusion model:
\begin{equation}
    x^t_0 = I^t - \tilde{I}^{t-1}, \ \ \tilde{I}^{t-1} = \mathcal{E}2\mathcal{V}(V^{t-1}),
\end{equation}
% \begin{equation}
%     \tilde{I}^{t-1} = \mathcal{E}2\mathcal{V}(V^{t-1}),
% \end{equation}
where $I^t$ represents the true intensity image of the scene at time $t$, $\tilde{I}^{t-1}$ represents the estimated intensity image of the scene at time $t-1$, and $\mathcal{E}2\mathcal{V}(\cdot)$ denotes the initial intensity predictor.
{From another perspective, the event voxel could be theoretically approximated as a degradation emanating from the temporal-domain residual image. Consequently, our framework embarks on a task analogous to image restoration, thereby harnessing the full generative capability of DDPM.}

In terms of generating target residual image $x^t_0$ under certain conditions using a diffusion model, the forward Markovian diffusion process $q$ adds Gaussian noise to the data at intervals, similar to DDPM~\cite{DDPM}:
\begin{equation}
    q(x^t_{\tau}|x^t_{\tau-1}) = \mathcal{N}(x^t_{\tau}; \sqrt{1-\beta_{\tau}} x^t_{\tau-1}, \beta_{\tau}\mathbf{I}),
\end{equation}
where $\beta_{\tau} \in (0, 1)$ for all $\tau = 1,...,\mathcal{T}$. $\beta_{\tau}$ are pre-chosen hyperparameters which determine the variance of the noise added at each iteration.
$\mathcal{T}$ is the number of steps in the iteration.
$\mathbf{I}$ represents the identity matrix.
Incorporating intermediate steps, we can obtain the distribution of $x^t_{\tau}$ given $x^t_{0}$, simplifying the noise addition:
\begin{equation}
    q(x^t_{\tau}|x^t_{0}) = \mathcal{N}(x^t_{\tau}; \sqrt{\Bar{\alpha}_{\tau}}x^t_{0}, (1-\Bar{\alpha}_{\tau})\mathbf{I}),
\end{equation}
where $\alpha_{\tau} \coloneqq 1 - \beta_{\tau}$ and $\Bar{\alpha}_{\tau} \coloneqq \prod^{\tau}_{i=1}\alpha_{i}$.
In order to extract high-frequency information from short-term events, we use the $V^t$, $\tilde{I}^{t-1}$, and the intermediate states of the ConvLSTM $s^{t-1}$ as conditioning inputs to train the diffusion model.
The pseudo-code for training is shown in Alg.~\ref{alg:train}.

The diffusion model generates content through a step-by-step denoising process executed in a reverse  Markov chain.
According to Bayesian theory, conditional probabilities can be derived as follows:
\begin{equation}
    p_{\theta}(x^t_{\tau-1}|x^t_{\tau}, V^t,\tilde{I}^{t-1},s^{t-1}) =  \mathcal{N}(x^t_{\tau-1};\mu_{\theta}(x^t_{\tau}, V^t,\tilde{I}^{t-1},s^{t-1},\tau),\sigma^2_{\tau}\mathbf{I}),
\end{equation}
\begin{equation}
    \mu_{\theta}(x^t_{\tau}, V^t,\tilde{I}^{t-1},s^{t-1},\tau) = \frac{1}{\sqrt{\alpha_{\tau}}}(x^t_{\tau} - \frac{\beta_{\tau}}{\sqrt{1-\Bar{\alpha}_{\tau}}}\epsilon_{\theta}(x^t_{\tau},V^t,\tilde{I}^{t-1},s^{t-1},\tau)),
\end{equation}
where $\sigma^2_{\tau} = \frac{1-\Bar{\alpha}_{\tau-1}}{1-\Bar{\alpha}_{\tau}} \beta_{\tau}$, $\theta$ denotes model parameters which are optimized by maximizing the variational lower bound (VLB) during the training phase.
In accordance with this conditional probability distribution, we can progressively generate the predicted intensity image at the sampling stage, and subsequently obtain the video as demonstrated in Alg.~\ref{alg:sample} and Fig.~\ref{fig:samplingAca}.

\begin{multicols}{2}
\begin{algorithm}[H]
\scriptsize  
\caption{Training on a scene}\label{alg:train}
\KwIn{A set of events voxel grids $\{V^t\}^{T-1}_{t=0}$; A set of intensity images corresponding to events $\{I^t\}^{T-1}_{t=0}$; A initial intensity predictor $\mathcal{E}2\mathcal{V}$}

    $s^0 = None$
    
    \For{$t=1,...,T-1$}{
        $\tilde{I}^{t-1} = \mathcal{E}2\mathcal{V}(V^{t-1})$\;
        
        $x^t_0 = I^t - \tilde{I}^{t-1}$\;
        
        $\tau \sim Uniform(\{1,...,\mathcal{T}\})$\;

        $\epsilon \sim \mathcal{N}(0, I)$\;

        $x^t_{\tau} = \sqrt{\Bar{\alpha}_{\tau}}x^t_{0} + \sqrt{1-\Bar{\alpha_{\tau}}}\epsilon$\;

        Take gradient descent step on 
        $\nabla_{\theta} \Vert \epsilon - \epsilon_{\theta}(x^t_{\tau},V^t,\tilde{I}^{t-1},s^{t-1},\tau) \Vert_1$\;

        $s^{t} = s_{\theta}(x^t_{\tau},V^t,\tilde{I}^{t-1},s^{t-1},\tau)$\;
        
    }
    
\end{algorithm}
\begin{algorithm}[H]
\scriptsize  
\caption{Sampling on a scene}\label{alg:sample}
\KwIn{A set of events voxel grids $\{V^t\}^{T-1}_{t=0}$; A initial intensity predictor $\mathcal{E}2\mathcal{V}$; A trained denoising model with weight $\theta$ }
\KwResult{A predicted video $\{\tilde{I}^0, \hat{I}^{t}\}^{T-1}_{t=1}$}

    $s^0 = None$
    
    \For{$t=1,...,T-1$}{
        $\tilde{I}^{t-1} = \mathcal{E}2\mathcal{V}(V^{t-1})$\;
        
        $x^t_{\mathcal{T}} \sim \mathcal{N}(0, I)$\;

        \For{$ \tau = \mathcal{T} - 1,...,1 $}{
            \eIf{$\tau > 1$}{
                $z \sim \mathcal{N}(0, I)$\;
            }{
                $z = 0$\;
            }
            $x^t_{\tau-1} = \mu_{\theta} + \sqrt{\sigma_{\theta}}z$
        }
        
        $\hat{I}^t = \tilde{I}^{t-1} + x^t_{0}$\;

        $s^{t} = s_{\theta}(x^t_{\tau},V^t,\tilde{I}^{t-1},s^{t-1},\tau)$\;
        
    }
    
\end{algorithm}
\end{multicols}

\subsection{Triple-path Conditional Model Architecture}\label{sec:CMA}

\noindent\textbf{Low-Frequency Intensity Estimation.}
As mentioned in Sec.~\ref{sec:CDF}, we attempt to address gap of data distribution by modifying the diffusion objective to include the difference image in the time domain.
However, temporal residual guided diffusion without any other conditions relies too much on the intensity image at the previous time step, which is not quite accurate.
We feed the generated results from the previous moment of $\mathcal{E}2\mathcal{V}$ as the initial intensity estimation into the network, and eliminate the error of the pre-trained $\mathcal{E}2\mathcal{V}$ through the accumulation of events in time domain.
Specifically, we use a simple multi-scale convolutional layer to extract features of initial intensity estimation, as shown in the following equation:
\begin{equation}
    \mathcal{F}(\Tilde{I}^{t-1})_{l+1} = {\rm ConvBlock}_l(\mathcal{F}(\Tilde{I}^{t-1})_l),
\end{equation}
where $\mathcal{F}(\ast)_{l}$ represents the features at layer $l$. Each ConvBlock consists of two convolutional layers and a downsampling layer (except for the last one).

\noindent\textbf{Recurrent Encoder for Temporal Motion Information.}
To accurately estimate the low-frequency information of the scene and predict the scene brightness more realistically, it is necessary to fully accumulate event data in the time domain.
We use multi-scale ConvLSTM~\cite{ConvLSTM} to extract long-term and short-term features of events voxels:
\begin{small}
\begin{equation}
    \mathcal{F}(V^t)_{l+1} = {\rm ConvLSTM}_l(\mathcal{F}(V^t)_l, s^{t-1}),
\end{equation}
\end{small}
where $s^{t-1}$ represents the hidden features of the events voxel from the previous time step.
\begin{figure}[t]
    \centering
    \includegraphics[width=\linewidth, angle=0]{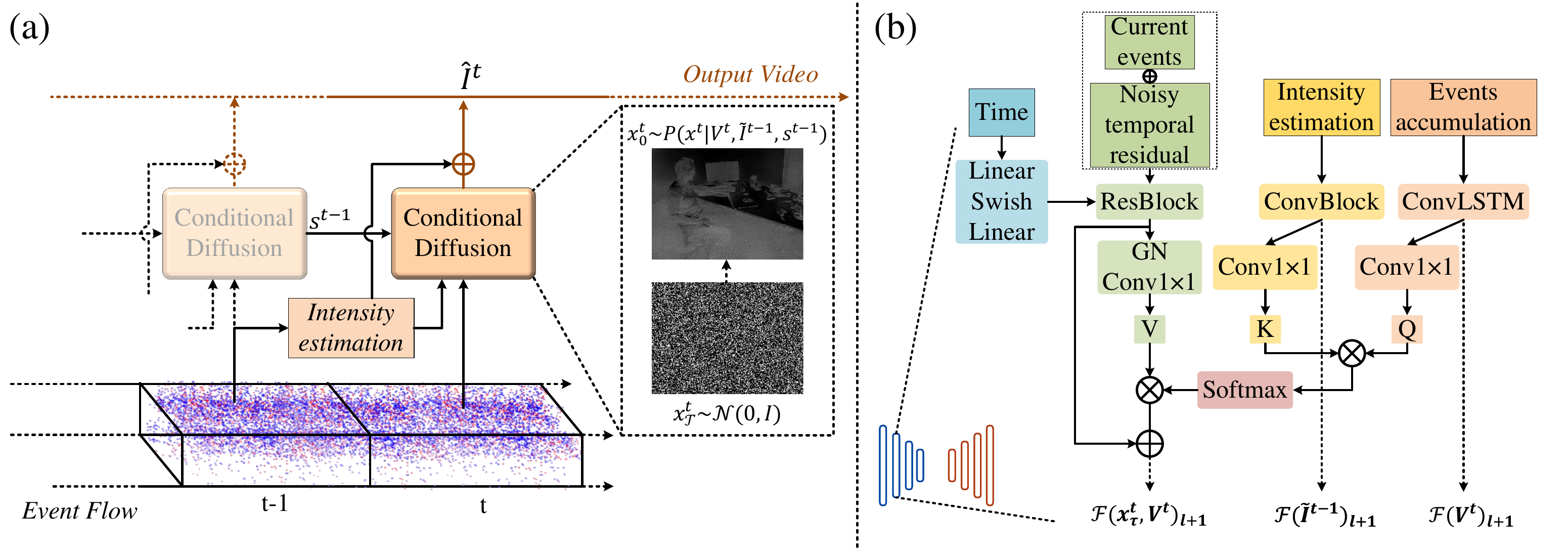}
    \caption{(a). Overview of sampling a video. The conditional diffusion model utilizes current events, intensity estimation, and features from accumulated events in the previous moment as guidance. This process generates high-frequency temporal residuals, contributing to the intensity image for each frame when added to the initial intensity estimation. (b). Overview of ResBlock with Cross Attention. Focus on events accumulation and intensity estimation features on the noisy temporal residuals, where GN denotes group normalization.}
    \label{fig:samplingAca}
\end{figure}

\noindent\textbf{Attention-based High-Frequency Prior Enhancement.}
We concatenate events voxel $V^t$ with the noisy intensity differential image $x^{t}_{\tau}$, both serving as representations of scene brightness variations physically, and input them into the same encoder to generate high-frequency details.
The encoder feature extraction process for this path can be expressed by the following formula:
\begin{small}
\begin{equation}
    \mathcal{F}(x^t_{\tau}, V^t)_{l+1} = {\rm CAtt}_l(\mathcal{F}(x^t_{\tau}, V^t)_l, \mathcal{F}(V^t)_{l}, \mathcal{F}(\Tilde{I}^{t-1})_{l}).
\end{equation}
\end{small}

Except for the highest resolution scale, where features are extracted using ResBlock, all other scales utilize ResBlock with cross attention as illustrated in Fig.~\ref{fig:samplingAca}.
In order to fully exploit the low-frequency features of temporally accumulated information and initial intensity estimation, we employ a cross-attention mechanism between three encoders.
Specifically, performing linear mapping on three different features separately using $1\times1$ convolutions without bias, resulting in $Q,K,V$:
\begin{equation}
    Q = {\rm Conv}_{1\times1}( \mathcal{F}(V^t)_{l}),  K = {\rm Conv}_{1\times1}( \mathcal{F}(\Tilde{I}^{t-1})_{l}), V = {\rm Conv}_{1\times1}( \mathcal{F}(x^t_{\tau}, V^t)_l).
\end{equation}

Then, we calculate attention~\cite{SelfAtt, VIT} across encoders as shown in the following equation and add it to the original feature $\mathcal{F}(x^t_{\tau}, V^t)_l$:
\begin{equation}
    att = {\rm Softmax}(\frac{QK^T}{\sqrt{d_k}})V,
    \label{eq:attention}
\end{equation}
where $d_k$ denotes the number of channels of $Q$.

We utilize residual blocks with self-attention to aggregate low-level features from three encoders.
Skip connections and upsampling are then employed for decoding the features to obtain the final reconstruction.
For specific details, please refer to the supplementary materials.

\section{Experiments}
\begin{figure*}[t]
    \centering
    \includegraphics[width=1\linewidth, angle=0]{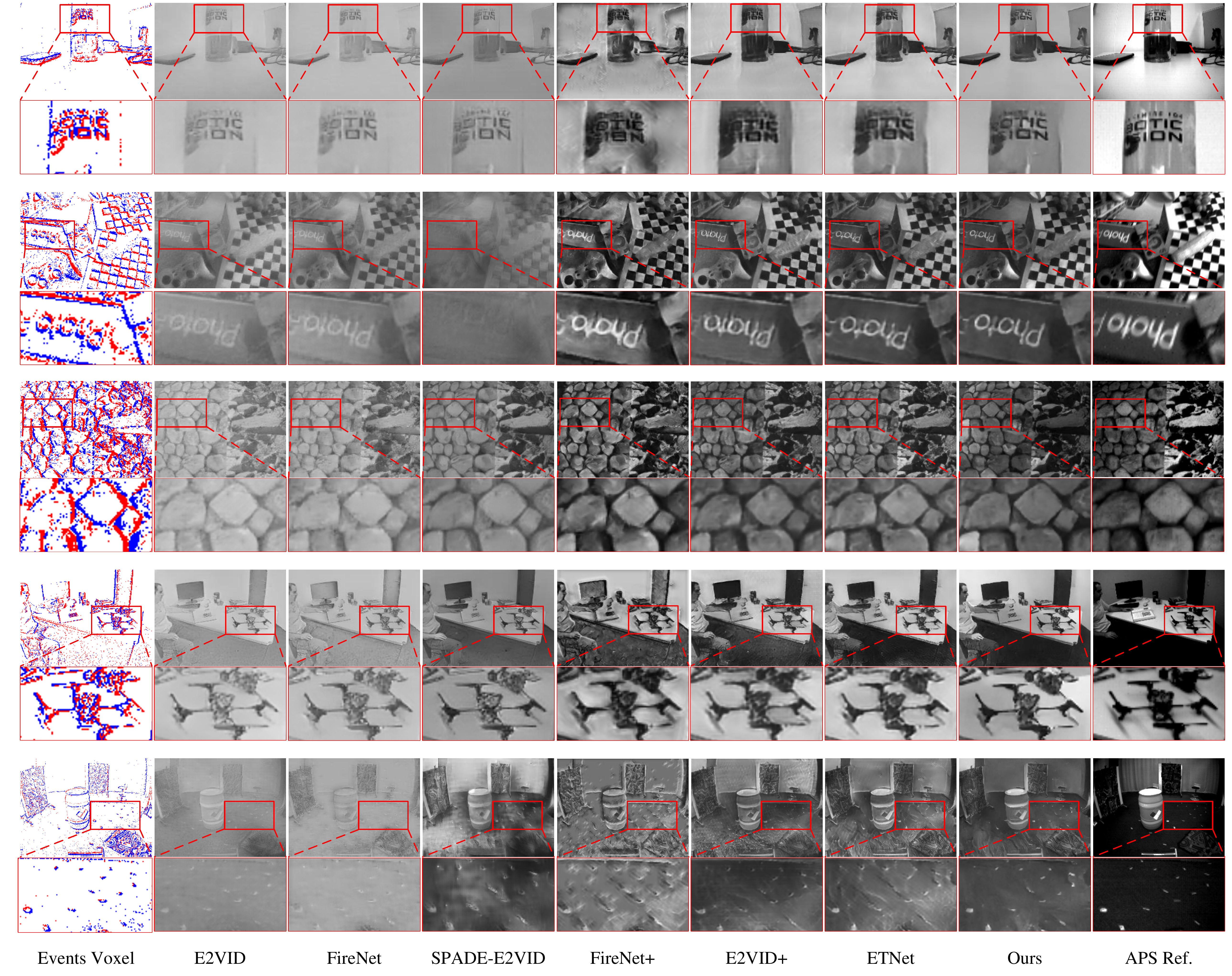}
    % \vspace{-3mm}
    \caption{Qualitative comparison on HQF (Row 1$\&$2), IJRR (Row 3$\&$4), and MVSEC (Row 5). Our results possess brightness distributions closest to the APS reference frame and the sharpest texture details.}
    \label{fig:comparison}
\end{figure*}

{
\setlength{\tabcolsep}{2.1pt}
\begin{table*}[t]
    \scriptsize 
    \centering
        \caption{Quantitative evaluation on multiple real datasets. Best in bold.}
        
        \begin{tabular}{c | c c c | c c c | c c c }
         \hline
        \multirow{2}{*}{Methods} & \multicolumn{3}{c|}{IJRR~\cite{IJRR}} & \multicolumn{3}{c}{HQF~\cite{HQF}} & \multicolumn{3}{|c}{MVSEC~\cite{MVSEC}} \\
         \cline{2-10}
          & MSE$\downarrow$ & SSIM$\uparrow$ & LPIPS$\downarrow$ & MSE$\downarrow$ & SSIM$\uparrow$ & LPIPS$\downarrow$ & MSE$\downarrow$ & SSIM$\uparrow$ & LPIPS$\downarrow$ \\
         \hline
        FireNet~\cite{FireNet}      & 0.1226              & 0.3448              & 0.2340
                                    & 0.0735              & 0.4035              & 0.2495   
                                    & 0.1613              & 0.2355              & 0.3189\\
                                    
        E2VID~\cite{E2VID}          & 0.1291              & 0.3490              & 0.2140  
                                    & 0.0644              & 0.4157              & 0.2288
                                    & 0.1643              & 0.2496              & 0.2900\\
                       
        SPADE-E2VID~\cite{SPADE}    & 0.0773              & 0.3696              & 0.2142
                                    & 0.0648              & 0.3564              & 0.2683     
                                    & 0.0994              & 0.2759              & 0.2936\\
        
        FireNet+~\cite{HQF}         & 0.0954              & 0.3011              & 0.2547
                                    & 0.0512              & 0.3382              & 0.2900     
                                    & 0.1271              & 0.2088              & 0.3843\\
                                    
        E2VID+~\cite{HQF}           & 0.0646              & 0.3688              & 0.1929
                                    & \textbf{0.0363}     & 0.4327              & 0.2160     
                                    & 0.0817              & 0.2762              & 0.2823\\
                      
        ETNet~\cite{ETNet}          & 0.0694              & 0.3673              & 0.1863
                                    & 0.0405              & 0.4004              & 0.2269   
                                    & 0.0889              & 0.2581              & 0.3016\\
        \hline
        Ours                        & \textbf{0.0597}     & \textbf{0.3856}     & \textbf{0.1694}
                                    & 0.0441              & \textbf{0.4496}     & \textbf{0.2039}
                                    & \textbf{0.0715}     & \textbf{0.2912}     & \textbf{0.2789}\\
          \hline
        \end{tabular}

    \label{tab:metrics}
\end{table*}
}

\subsection{Experimental Setup}

During training, we use the same dataset as ~\cite{E2VID-TPAMI}, which generates events on the COCO ~\cite{COCO} dataset using the event simulator ESIM~\cite{ESIM}.
An alternative simulated dataset~\cite{HQF} could showcase scenes with a wider dynamic range and a threshold distribution more akin to existing datasets like IJRR and MVSEC. This similarity might bolster the performance of reconstruction models such as E2VID+~\cite{HQF} and FireNet+~\cite{HQF}. Nevertheless, it may not be suitable for training our temporal residual framework. This constraint is linked to the inclusion of masked images in the dataset, aimed at simulating foreground and background motion. This methodology may introduce occlusion events, reminiscent of the occlusion challenges encountered in optical flow tasks.

In our experiments, to confirm the generalization performance of our method on real-world events, we use multiple real datasets to test, including IJRR~\cite{IJRR}, HQF~\cite{HQF} and MVSEC~\cite{MVSEC}.
The iteration number $\mathcal{T}$ takes the value of 2,000 during the training and testing phases.
We use ETNet~\cite{ETNet} as the initial intensity predictor.
Then, we conduct an ablation study to find out the effectiveness of different strategies.
Please refer to the supplementary materials to find more details about the experiments.

\begin{figure*}[t]
    % \vspace{-3mm}
    \centering
    \includegraphics[width=1\linewidth, angle=0]{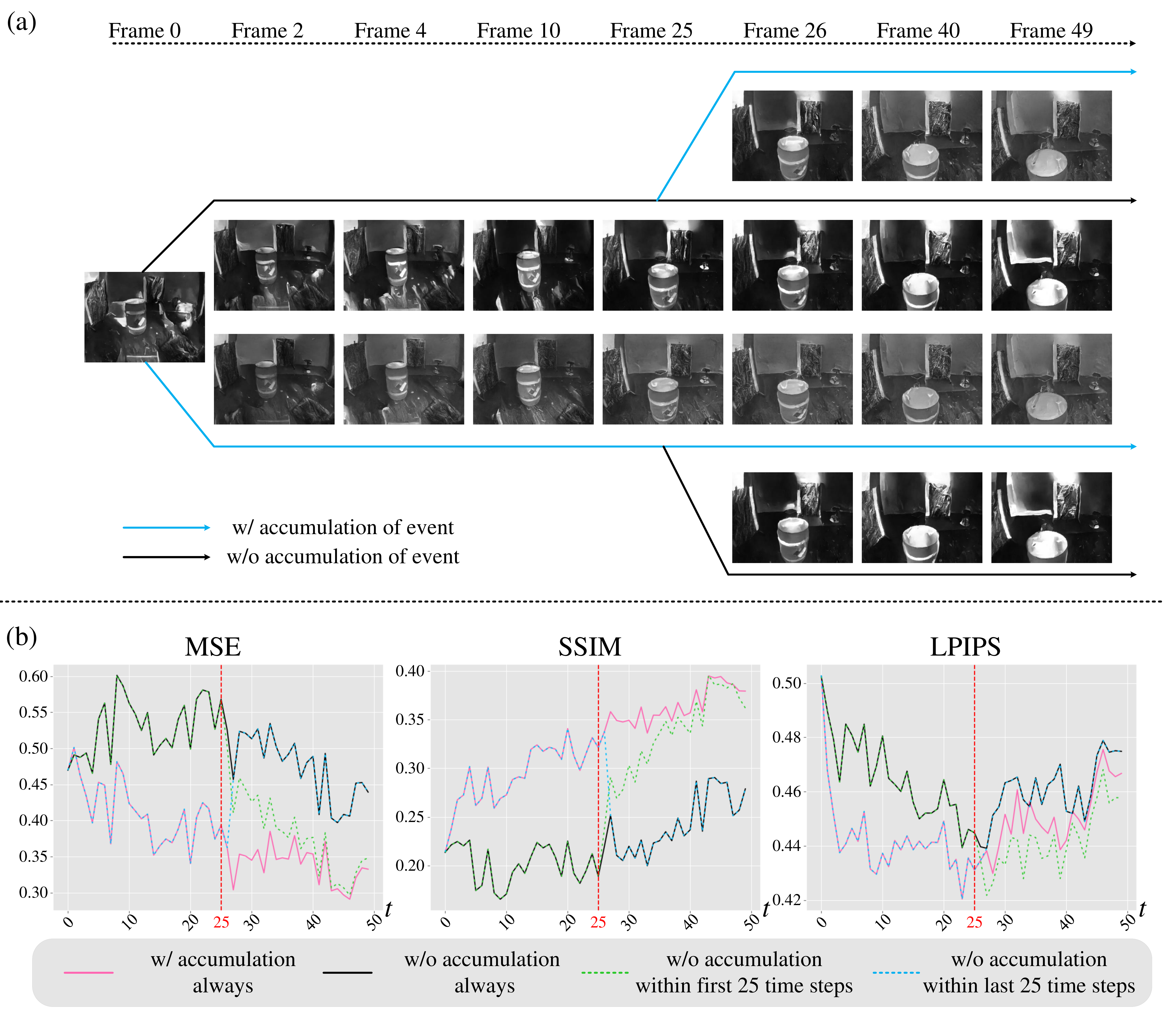}
    
    \caption{(a) Visual comparison between w/o and w/ event temporal accumulation. (b) Quantitative results of (a). We test different event accumulation ways across multiple metrics over time (Best viewed with zoom in). `w/o accumulation within first 25 time steps' represents the process in the first row of (a). `w/o accumulation within last 25 time steps' represents the process in the fourth row of (a).}
    \label{fig:recurrent}
    %\vspace{-3mm}
\end{figure*}
\subsection{Comparison with the State-of-the-Art Methods}

We compare the proposed method against various event-based reconstruction methods, including E2VID~\cite{E2VID-TPAMI}, E2VID+~\cite{HQF}, FireNet~\cite{FireNet}, FireNet+~\cite{HQF}, SPADE-E2VID~\cite{SPADE} and ETNet~\cite{ETNet}.
The reconstructed images generated by each method are compared with the APS reference frames, and the mean square error (MSE), structural similarity (SSIM~\cite{SSIM}) and perceptual loss (LPIPS~\cite{LPIPS}) indicators are used to quantitatively evaluate.
We perform tests on all sequences of each dataset, Table~\ref{tab:metrics} shows the comparison of average metrics results for each dataset, and Fig.~\ref{fig:comparison} shows a visual comparison on some of the sequences.
Of note, it is essential to recognize that the APS reference frames within the existing real dataset are captured by an integral camera operating within the same optical path.
Due to limitations in dynamic range and differences in sampling mechanisms, there exists a considerable gap of the brightness distribution of the scene as described between APS frame and the intensity image based on event reconstruction.
To ensure experimental fairness, all reconstructed images and APS reference frames undergo histogram equalization before evaluation like SPADE-E2VID~\cite{SPADE}.

Our approach has achieved state-of-the-art performance across almost all metrics.
In particular, attributed to a thorough exploration of the high-frequency priors inherent in events, our results with superior structural details, exhibiting the highest SSIM metrics across all datasets, achieving improvements of 4.3$\%$, 3.9$\%$ and 5.4$\%$ over the second-best on IJRR, HQF and MVSEC, respectively.
In terms of visual comparison, our approach emphasizes detailed reconstruction, as evidenced by the clear textual outlines in the first and second rows of Fig.~\ref{fig:comparison}, distinctly separated from the background.
The other methods either suffer from low contrast or lack completeness in reconstructing details, often containing artifacts and blur.

\subsection{Ablation Study}

To validate the rationality of the various strategies proposed, we conduct multiple sets of ablation experiments.
The iterations for different models are kept identical and tested on the MVSEC dataset for experimental consistency.
Given the sparse event flow, it presents greater challenges for reconstruction.

\noindent\textbf{Effect of Temporal Residual Diffusion.}
In line with Sec.~\ref{sec:CDF}, due to the disparate modalities between event flow and intensity images, employing event data as a condition to guide the diffusion model for directly generating reconstructed images is theoretically unfeasible.
In keeping all other conditions constant, we alter the optimization target of the diffusion model to intensity image, and the quantitative results of the testing are shown in the first row of Table \ref{tab:ablation}.
Please refer to supplementary materials for further visual comparison.
Due to variations in the data distribution of the events and light intensity, directly generating intensity image results in a lack of sufficient texture details, especially in areas with dimly lit scenes.

\noindent\textbf{Effect of Cross Encoder Attention Mechanism.}
In Sec.~\ref{sec:CMA}, we mentioned the use of a cross attention mechanism to focus on features across different encoders.
We propose this strategy aiming to utilize temporal event features for correcting the errors from intensity estimation.
The method that does not employ cross-attention mechanisms during the feature extraction stage has been retrained, and results are presented in the third row of Table \ref{tab:ablation}.
The visual comparison indicates that the absence of cross-attention mechanisms results in lower contrast.
The brightness described differs significantly from the real scene, albeit with good reconstruction of edge details.

\begin{figure}[t]
    \centering
    \includegraphics[width=1.0\linewidth, angle=0]{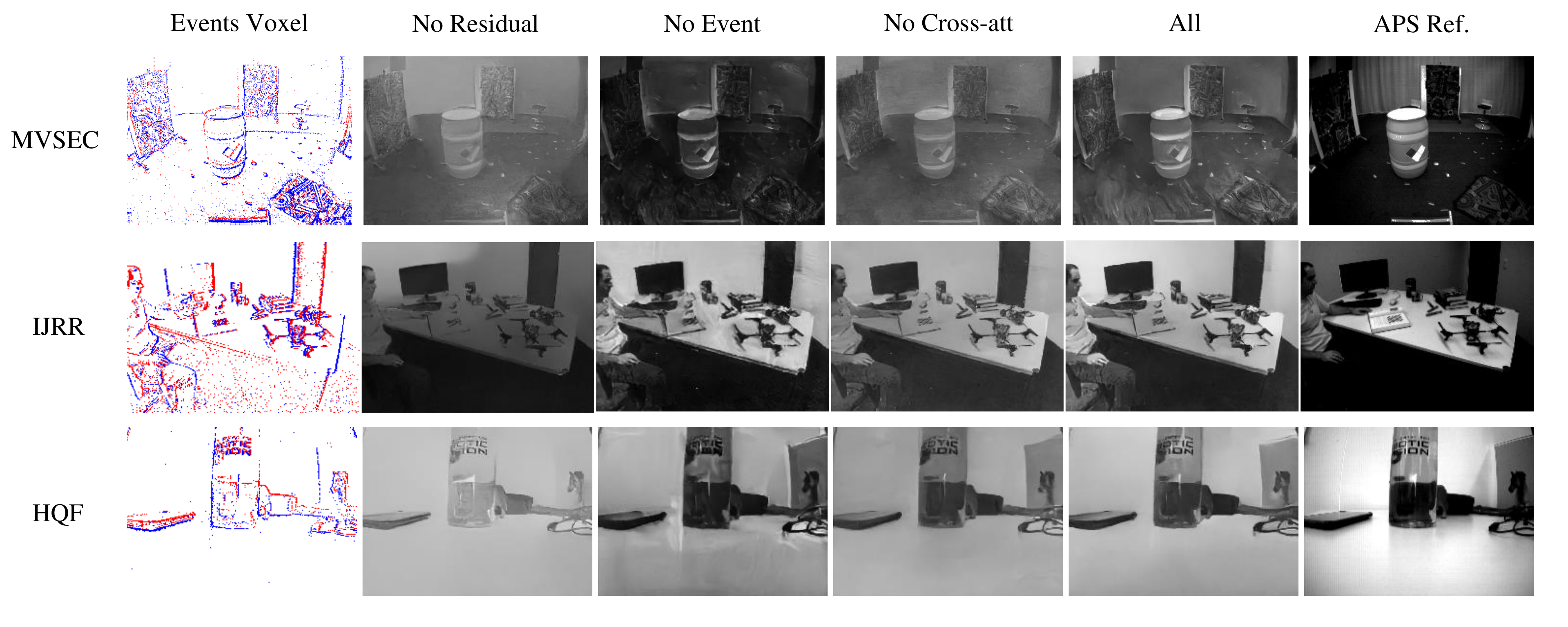}
    \caption{The visual comparison of ablation experiments. `No residual' in the second col signifies that the diffusion model generates target intensity image directly rather than temporal residual. `No event' in the third col denotes the usage of initial intensity estimation as a only condition to guide the diffusion model. `No cross-att' in the fourth col indicates the absence of cross-encoder attention mechanisms during the encoding phase. `All' in the fifth col represents the results of our final model.}
    \label{fig:cmpablation}
\end{figure}
\begin{figure}[t]
    \centering
    \includegraphics[width=1.0\linewidth, angle=0]{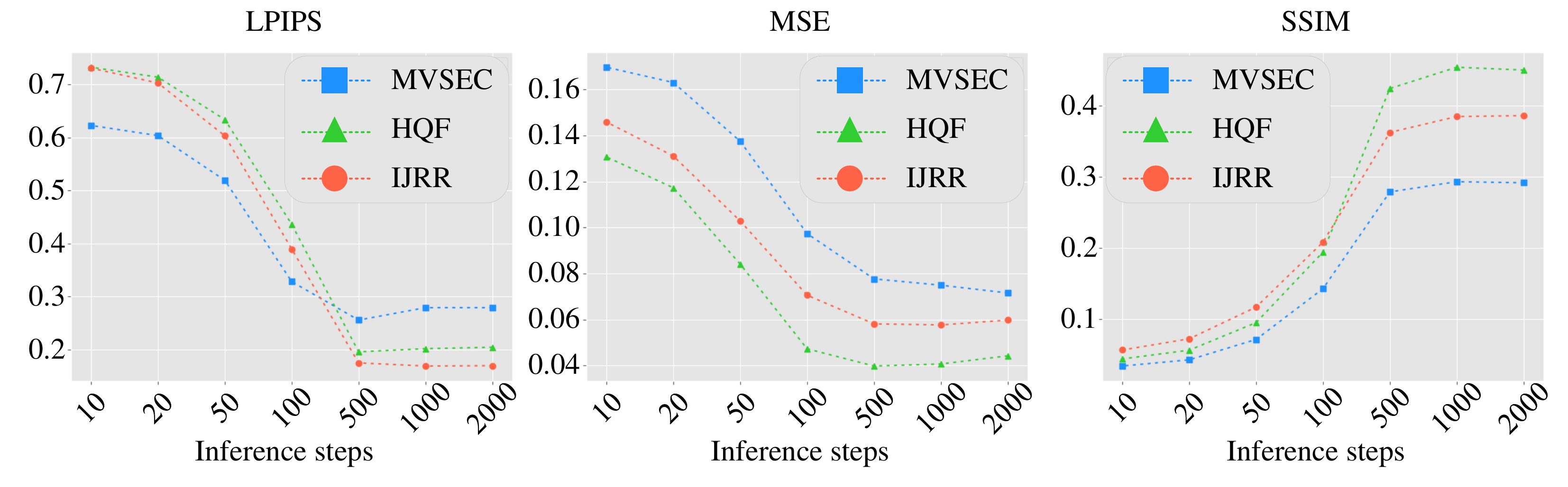}
    \caption{Comparison with different inference steps on multiple datasets.}
    \label{fig:cmpinfer}
\end{figure}
% }
\noindent\textbf{Effect of Recurrent Encoder for Event Accumulation.}
In line with Sec.~\ref{sec:CMA}, we employ ConvLSTM to aggregate temporal information for more accurate intensity estimation.
Fig.~\ref{fig:recurrent} presents a comparison of whether temporal event features are utilized at different time steps.
Furthermore, the lack of accumulated events indicates that the hidden state and cell state of the ConvLSTM input remains empty.
The two middle rows depict the results generated respectively without using temporal cumulative event features at all and with complete usage always.
In the second row, the overall brightness between each frame remains nearly constant, yet there are significant local brightness differences.
This indicates the unreliability of relying solely on input intensity estimation, as there exists a considerable gap between it and the actual intensity.
The third row demonstrates the accumulation of temporal events voxel.
Here, the global brightness gradually changes, and areas initially overexposed or underexposed tend towards moderate brightness, validating the theory behind our design of the temporal recurrent encoder for correcting intensity errors.
The first row, beginning the accumulation of events at a midpoint, gradually enhances contrast in the reconstructed image, and the corresponding metrics tend to approximate the results of continuous event accumulation.
In the meantime, in the fourth row, abruptly ceasing to use the preceding temporal features at the midpoint causes a sudden deterioration in the reconstructed results, with the corresponding metrics plummeting to the level of non-cumulative event results across the entire duration.
The quantitative results shown in subplot (b) of Fig.~\ref{fig:recurrent} also confirm the above analysis.

\noindent\textbf{Effect of Event-based Conditional Diffusion.}
Our proposed method aims to exploit the high-frequency information inherent in sparse events to enhance texture details, as demonstrated in our comparative experiments where our results yield sharper edge regions. 
We retrain a diffusion model guided solely by initial intensity estimates, and its results are shown in the fourth row of Table~\ref{tab:ablation}. 
Visual assessments indicate a notably poor reconstruction quality, struggling not only to approximate the real scene's brightness distribution but also exhibiting blurriness, artifacts, and a lack of sufficient texture details at local regions.
{
\setlength{\tabcolsep}{3.8pt}
\begin{table}[t]
\footnotesize  
    \centering
        \caption{Quantitative comparison of different strategies. Best in bold.}
        \begin{tabular}{c c c c | c c c  }
         \hline
        \multicolumn{4}{c|}{Strategies}                          & \multicolumn{3}{c}{MVSEC} \\
        \hline
        \scriptsize{Residual} & \scriptsize{Recurrent} & \scriptsize{Cross-att} & \scriptsize{Event} & \scriptsize{MSE$\downarrow$} & \scriptsize{SSIM$\uparrow$} & \scriptsize{LPIPS$\downarrow$} \\

                            &$\checkmark$       & $\checkmark$      & $\checkmark$      &0.0902             &0.2565             &0.2977\\
                       
        $\checkmark$        &                   & $\checkmark$      & $\checkmark$      &0.1021             &0.2472             &0.3023\\
             
        $\checkmark$        & $\checkmark$      &                   & $\checkmark$      &0.0812             &0.2653             &0.2958\\
                      
        $\checkmark$        &                   & $\checkmark$      &                   &0.0960             &0.2309             &0.3189\\
        \hline
        $\checkmark$        & $\checkmark$      & $\checkmark$      & $\checkmark$      &\textbf{0.0707}    &\textbf{0.2768}    &\textbf{0.2892}\\  \hline

        \end{tabular}
        % \vspace{-3mm}

    \label{tab:ablation}
\end{table}
}

\noindent\textbf{Effect of Different Inference Steps.}
\cref{fig:cmpinfer} presents the metric results with varying inference steps across different datasets.
It can be observed that with an increase in noise step size, the metrics gradually improve, especially the SSIM and LPIPS indicators show a notable enhancement.
Due to the MSE metric's greater emphasis on brightness differences between the image and its reference, while the noise step focuses on generating details, convergence is notably rapid for the MSE metric.
Considering all measurement results, convergence is approached when the iterations exceed 1,000.

\section{Conclusion}
The advantage of event-based video reconstruction lies in high dynamic range and rapid motion capture.
However, prevailing methods overly prioritize temporal information, resulting in over-smoothing and blurry artifacts.
Our solution, the temporal residual guided diffusion Framework, adeptly integrates temporal features, low-frequency texture, and high-frequency event features.
Three key conditioning modules enhance the Denoising Diffusion Probabilistic Model, ensuring accurate reconstructions.
By leveraging the temporal-domain residual features, our model captures both temporal and high frequency event information.
Our framework excels in mitigating over-smoothing and artifacts, evident in extensive benchmark experiments, surpassing prior methods.
This marks a significant stride toward high-quality event-based video reconstruction, addressing persistent challenges in the field.

\noindent\textbf{Acknowledgement.}
This work is partially supported by National Natural Science Foundation of China under Grant No.62302041, 62322204, 62131003, China National Postdoctoral Program under contract No.BX20230469, and Beijing Institute of Technology Research Fund Program for Young Scholars.

\bibliographystyle{splncs04}
\bibliography{main}
\end{document}